\documentclass{article}
\usepackage{spconf,amsmath,graphicx,amssymb}
\usepackage{algorithm, algcompatible}
\usepackage{lipsum}

\usepackage{graphicx}
\usepackage{colortbl,array} 
\usepackage{multirow,bigdelim}
\usepackage{booktabs}

\algnewcommand\INPUT{\item[\textbf{Input:}]}
\algnewcommand\OUTPUT{\item[\textbf{Output:}]}

\title{Blind Image Deconvolution using Student's-t Prior \\ with Overlapping Group Sparsity}
\name{In S. Jeon \qquad Deokyoung Kang \qquad Suk I. Yoo}
\address{Dept. of Computer Science and Engineering, Seoul National University, Seoul, South Korea} 

\begin{document}
%
\maketitle
\begin{abstract}

In this paper, we solve blind image deconvolution problem that is to remove blurs form a signal degraded image without any knowledge of the blur kernel. Since the problem is ill-posed, an image prior plays a significant role in accurate blind deconvolution. Traditional image prior assumes coefficients in filtered domains are sparse. However, it is assumed here that there exist additional structures over the sparse coefficients. Accordingly, we propose new problem formulation for the blind image deconvolution, which utilize the structural information by coupling Student's-t image prior with overlapping group sparsity. The proposed method resulted in an effective blind deconvolution algorithm that outperforms other state-of-the-art algorithms.

\end{abstract}
\section{Introduction}
\label{sect:intro} 

Image deconvolution is the problem of restoring an image $x$ from its blurred and noisy version $y$. Generally, the image $y$ is modeled as 
\begin{equation} \label{eq:1}
    y = h \otimes  x + \varepsilon  \, ,
\end{equation}
where $\otimes$ denotes two-dimensional convolution, $h$ is blur kernel, and $\varepsilon $ is noise term. Since there exists infinitely many solution for $x$, (\ref{eq:1}) is an ill-posed problem \cite{ill1,ill2}. Hence, a regularization reflecting our prior knowledge is necessary to be imposed on the image $x$ to obtain a meaningful solution. The regularization is generally embedded by assigning prior distribution $p(x)$ in the Bayesian formulation of the problem. 

The choice of the image prior is varying, but the most popular one is the sparsity-enforcing prior. It is well-known that when high-pass filters are applied to natural images, the resulting coefficients are sparse \cite{natural1,natural2}; i.e., most of the coefficients are zero or very small while only a small number of them are large (e.g., at the edges). This important characteristic has been utilized in many image deconvolution algorithms. Fergus et al. \cite{fergus:2006} introduced a mixture-of-Gaussians (MoG) prior with a filtered image representation and showed the proposed approach was practical. After the success of his work, many researchers have subsequently suggested other kinds of the sparsity-enforcing image prior such as total variation \cite{shan:2008,cho:2009}, hyper-Laplacian \cite{ker:2009}, and Student's-t \cite{dim:2009}.

When the blur kernel $h$ in (\ref{eq:1}) is unknown in addition to the unknown image $x$, it becomes a ``blind'' image deconvolution problem. This is much difficult to solve than the non-blind image deconvolution problem since there exist infinitely many possible combinations of $x$ and $h$ making it severely ill-posed. Levin et al. \cite{levin:2009} reveals that the conventional sparsity-enforcing priors \cite{fergus:2006,shan:2008,cho:2009,ker:2009} confront a limitation in the case of blind image deconvolution; they indicate that these priors flavor the blurred image over the correct one because the blurred image often has more zero coefficients (i.e. more sparse) than the clear image in high-pass filtered domains. 

To avoid this ``no-blur'' solution, Levin et al. \cite{levin:2011} employ a marginal likelihood optimization. Krishnan et al. \cite{ker:2011} suggest a normalized sparsity measure to mitigate this problem. Moreover, Wipf and Zhang \cite{wipf:2014} emphasize the power of image prior that can discriminate a good sharp image from blurred images is critical to recovering the unknown image. Both Babacan \cite{babacan:2012} and Perrone \cite{perrone:2015} are good examples of this approach. They \cite{wipf:2014,babacan:2012,perrone:2015} all utilized non-convex image priors to strongly promote the sparsity of the signal, regardless of the natural statistics, and achieved state-of-the-art blind image deconvolution result. 

To overcome the limitation of the traditional image priors, there is another growing interest for structured sparsity \cite{peyr:2011,figu:2011,bach:2012,sele:2013}. In fact, all the image priors presented in \cite{fergus:2006,shan:2008,cho:2009,ker:2009,dim:2009,levin:2009,levin:2011,ker:2011,wipf:2014,babacan:2012,perrone:2015} assumes the coefficient in the filtered domains is sparse; however, large values of the coefficients generally do not occur in isolation. Hence authors in \cite{peyr:2011,figu:2011,bach:2012,sele:2013} assumes the coefficients exhibits a simple form of structure that is called structured sparsity. Liu et al. \cite{liu:2015} adapt the group sparsity regularizer to recover a noise corrupted image, and it is proven to be very effective in alleviating staircase effects. Shi et al. \cite{shi:2016} also shows a hyper-Laplacian constrained with overlapping group sparsity leads to a good image deconvolution result. 

In this paper, we utilize the structured sparsity \cite{peyr:2011,figu:2011,bach:2012,sele:2013} to solve the blind image deconvolution problem. We presented new problem formulation for the blind image deconvolution, which incorporates conventional Student's-t image prior \cite{dim:2009} with overlapping group sparsity regularizer \cite{sele:2013}. The proposed problem formulation resulted in an effective blind image deconvolution algorithm that outperforms recently introduced state-of-the-art algorithms such as \cite{levin:2011,babacan:2012,perrone:2015}. We verified this result in the experimental section.

The rest of this paper is organized as follows. Section 2 describes our modeling of the blind deconvolution problem. The inference algorithm is outlined in section 3. In Section 4 we present the experimental results. Section 5 concludes the paper.

\section{Problem Formulation}

Since we are in the discrete domain, convolution in (\ref{eq:1}) is equivalent to vector-matrix multiplication. Then we can rewrite the model as

\begin{equation} \label{eq:2}
    y = Hx + \varepsilon = Xh + \varepsilon \, .
\end{equation}
$y$, $x$, and $\varepsilon$ are lexicographically arranged $n^2$ dimensional vector. $H$ and $X$ are the two-dimensional convolution matrices obtained from the kernel $h \in \mathbb{R}^{k^2}$ and image $x \in \mathbb{R}^{n^2}$ respectively. We further assume the noise $\varepsilon$ is i.i.d Gaussian with variance $\sigma^2$, then we obtain following observation model:
\begin{equation} \label{eq:3}
  p(y|x,h) = \; \mathcal{N}(y|Hx, \sigma^2 I)\, .
\end{equation}

From Bayesian perspective, the goal of blind image deconvolution is to infer the unknown (latent) variables $x$ and $h$. Estimating these variables is usually done by maximizing a posterior distribution: $p(x,h|y) \propto p(y|x,h)p(x)p(y)$ where $p(y|x,h)$ is likelihood defined as (\ref{eq:3}). $p(x)$ and $p(h)$ are the prior distributions for the unknown image and the unknown blur kernel respectively. This approach is commonly referred to as the maximum a posterior (MAP) estimation.

\subsection{Kernel Prior}

For the kernel prior, we choose the flat prior:  $p(h) \propto 1$. Considering the second part (i.e. $y = Xh + \varepsilon$) in (\ref{eq:2}), we have $n^2$ observations and aim at estimating $k^2$ coefficients while solving the kernel $h$. Since the image size is usually much larger than the kernel size (i.e. $k^2 << n^2$), $n^2$ observations should be sufficient to obtain a good kernel estimate. Based on the fact, many authors \cite{wipf:2014,babacan:2012,perrone:2015} also used the flat prior on the kernel, enforcing only its non-negativity and normalization constraints.

\subsection{Student's-t Image Prior}

The image prior $p(x)$ is based on $m$ filtered versions of the image: $g_m = F_m x$, where $F_m$ are two dimensional convolution matrix obtained from high-pass filters: $\{f_m\}_{m=1}^M$. Specifically, we use the first-order differences between 4 local neighbors.

Assuming that each pixel $g_{m,i}$ at index $i$ follows Gaussian distribution with distinct precision $\gamma_{m,i}$, and the precision is Gamma random variable with the shape parameter $\alpha$ and the scale parameter $\beta$, we can define a hierarchical joint distribution for each $(g_{m,i},\gamma_{m,i})$ as follows: 
\begin{equation} \label{eq:4}
\begin{aligned}
    p(g_{m,i}, \gamma_{m,i}) &\propto p_{m,i}(g_{m,i}| \gamma_{m,i}) p(\gamma_{m,i}) \\ 
    &\propto \mathcal{N}(g_{m,i}|0,\gamma_{m,i}^{-1}) \textit{Gamma}(\gamma_{m,i} | \alpha,\beta) \;.
\end{aligned}
\end{equation}
The marginalization of (\ref{eq:4}) with respect to $r_{m,i}$ is equivalent to Student's-t distribution. Hence, this hierarchical prior closely resembles the Student's-t prior enforcing the sparseness of the image pixels in the filtered domains \cite{dim:2009}. Also, notice that new auxiliary variable $\gamma_{m,i}$ is introduced, which will be estimated jointly. 

By multiplying (\ref{eq:4}) across all the spatial indexes $i$ and the filter indexes $m$, we can obtain Student's t image prior as follows:
\begin{equation} \label{eq:5}
\begin{aligned}
    p(x,\gamma) & \propto p(x|\gamma)\,p(\gamma) \\
    & \propto \prod_{m=1}^{M}\mathcal{N}(x|0,(F_m^T \textit{diag} \{\gamma_m\} F_m)^{-1}) \\
    & \times \prod_{m=1}^M \prod_{i=1}^N  \textit{Gamma}(\gamma_{m,i} | \alpha,\beta)
\end{aligned}
\end{equation}
with $\gamma_m = (\gamma_{m,1}, \dots, \gamma_{m,N})^T$. 

MAP estimation is equivalent to minimizing the negative log posterior. Accordingly, with the equation (\ref{eq:3}), (\ref{eq:5}), and $p(h) \propto 1$, we get the following optimization problem for the blind image deconvolution:
\begin{equation} \label{eq:6}
\begin{aligned}
    & \min_{x,\gamma,h} -\log \, p(x,\gamma,h|y) \\
    = & \min_{x,\gamma,h} -\log(p(y|x,h))-\log(p(x,\gamma))-\log(p(h)) \\
    = &\min_{x,\gamma,h} \; ||Hx-y||^2 + \lambda_1 \, \psi(x,\gamma) \, ,
\end{aligned}
\end{equation}
where $\lambda_1 = 1/\sigma^2$ is a regularization parameter, and
\begin{equation} \label{eq:7}
\begin{aligned}
    \psi(x,\gamma) & = \sum_{m=1}^{M} x^T F_m^T \textit{diag} \{\gamma_m\}  F_m x \\
    & + \, 2 \sum_{m=1}^{M} \sum_{i=1}^{N} ((1-\alpha) \log \gamma_{m,i} + \beta \gamma_{m,i}).
\end{aligned}
\end{equation}
$\psi(x,\gamma)$ is a regularization term obtained from Student's-t image prior promoting the sparsity of coefficients in the filtered domains. However, it dose not take account the structural information among the coefficients.

\subsection{Overlapping Group Sparsity}

To capture the structural information among the coefficients, we define two-dimensional $W \times W$ points group in the two-dimensional signal $s$ as follows:
\begin{equation} \label{eq:8}
\begin{aligned}
    { \tilde { s }  }_{ (i,j),W }= \begin{bmatrix} { s }_{ (i-{ m }_{ 1 },j-{ m }_{ 1 }) } & \cdots  & { s }_{ (i-{ m }_{ 1 },j+{ m }_{ 2 }) } \\ { s }_{ (i-{ m }_{ 1 }+1,j-{ m }_{ 1 }) } & \cdots  & { s }_{ (i-{ m }_{ 1 }+1,j+{ m }_{ 2 }) } \\ \vdots  & \ddots  & \vdots  \\ { s }_{ (i+{ m }_{ 2 },j-{ m }_{ 1 }) } & \cdots  & { s }_{ (i+{ m }_{ 2 },j+{ m }_{ 2 }) } \end{bmatrix} \; \\ \in \mathbb{R}^{W \times W} \; ,
\end{aligned}
\end{equation}
with $m_1 = \left\lfloor \frac{W-1}{2} \right\rfloor, m_2 = \left\lfloor \frac{W}{2} \right\rfloor$, where $ \left\lfloor x \right\rfloor$ denotes the floor function, and $W$ is the window size. Hence, $\tilde{s}_{(i,j),W}$ is a group of $W \times W$ contiguous samples centered at $(i,j)$.

By stacking the columns of $ {\tilde{s}}_{(i,j),W}$, a vector $s_{(i,j),W}$ is obtained: $s_{(i,j),W}={\tilde{s}}_{(i,j),W}(:)$. Then, the overlapping group sparsity (OGS) functional \cite{liu:2015,shi:2016} is
\begin{equation} \label{eq:9}
\varphi_{\text{OGS}}(s) = \sum_{i,j=1}^{n} \left\| s_{(i,j),W} \right\|_2 \, . 
\end{equation}
$\varphi_{\text{OGS}}$ takes account all the overlapping groups of pixels on the spatial domain of the signal $s$. With (\ref{eq:9}), we define OGS regularization term for the blind deconvolution problem:
\begin{equation}\label{eq:10}
  \phi(x) = \sum_{m=1}^M \varphi_{\text{OGS}}(g_m) \; ,
\end{equation}
where $g_m = F_m x$. If $W=1$, $\phi$ is the commonly used anisotropic TV prior. In this sense, $\phi$ is also referred as generalized total variation \cite{sele:2015}.

Therefore, we present a novel problem formulation, which incorporates Student's-t image prior and the overlapping group sparsity, to solve the blind image deconvolution problem as follows:
\begin{equation} \label{eq:11}
    \min_{x,\gamma,h} \, \left( R(x) = ||Hx-y||^2 + \lambda_1 \psi(x,\gamma) + \lambda_2 \phi(x) \right)\\
\end{equation}
where the functional $\psi(x,\gamma)$ and $\phi(x)$ is given by (\ref{eq:7}) and (\ref{eq:10}) respectively. $\lambda_1$ and $\lambda_2$ are the regularization parameters. 

\section{Inference Algorithm}

 We use majorization-minimization (MM) as in \cite{liu:2015,sele:2013,sele:2015} to derive a computationally efficient algorithm to solve the problem (\ref{eq:11}). To find a majorizor of $R(x)$ in (\ref{eq:11}), we first find a majorizor of $\varphi_{\text{OGS}}(v)$ in (\ref{eq:9}). To this end, note that
\begin{equation} \label{eq:12}
\frac{1}{2 \left\| u \right\|_2 }  \left\| v \right\|_2^2 + \frac{1}{2} \left\| u \right\|_2 \ge \left\| v \right\|_2
\end{equation}
for all $v$ and $u\neq 0$ with equality when $u=v$. Substituting each group of $\varphi_{\text{OGS}}(v)$ into (\ref{eq:12}) and summing them, we get a majorizor of $\varphi_{\text{OGS}}(v)$ 
\begin{equation}
\begin{aligned}
&P(v,u) \\ 
&= \frac{1}{2} \sum_{i,j=1}^n \left( \frac{1}{\left\| u_{(i,j),W} \right\|_2}  \left\| v_{(i,j),W} \right\|_2^2 +  \left\| u_{(i,j),W} \right\|_2 \right)
\end{aligned}
\end{equation}
with
\begin{equation} \label{eq:14}
  P(v,u) \ge \varphi_{\text{OGS}}(v)\; , \; P(u,u) = \varphi_{\text{OGS}}(u)
\end{equation}
provided $ \left\| u_{(i,j),W} \right\|_2 \neq 0$ for all $i,j$. With a simple calculation, $P(v,u)$ can be rewritten as
\begin{equation} \label{eq:15}
 P(v,u) = \frac{1}{2} v^T \Lambda(u) v + C, 
\end{equation}
where C is constant that dose not depend on $v$, and $\Lambda(u)$ is a diagonal matrix with each diagonal component
\begin{equation} \label{eq:16}
[\Lambda(u)]_{l,l} = \sum_{i,j=-m_1}^{m_2} \left[ \sum_{w_1,w_2=-m_1}^{m_2} | u_{(r-i+k_1,t-j+k_2)} |^2 \right]^{-\frac{1}{2}}
\end{equation}
with $l=(t-1)n+r$, and $r,t=1,2,\dots,n$.

Substituting each $\varphi_{\text{OGS}}(g_m)$ in (\ref{eq:10}) into (\ref{eq:14}), a majorizor of $R(x)$ in (\ref{eq:11}) can be obtained by
\begin{equation} \label{eq:17}
\begin{aligned}
    G(x,x') & = ||Hx-y||^2 + \lambda_1 \psi(x,\gamma) + \lambda_2 \phi'(x,x') \\
    \ge R(x) &= ||Hx-y||^2 + \lambda_1 \psi(x,\gamma) + \lambda_2 \phi(x)\,,
\end{aligned}
\end{equation}
where
\begin{equation}
\begin{aligned}
    & \phi'(x,x') \\
    & = \sum_{m=1}^M P(g_m, g'_m) = \sum_{m=1}^M  x^T F_m^T \Lambda(F_m x') F_m x,
\end{aligned}
\end{equation}
where $x'$ is the estimation of $x$ at the previous iteration. 

MM algorithm solve the problem (\ref{eq:11}) by iteratively minimizing $G(x,x')$ in (\ref{eq:17}). Since the first term in $G(x,x')$ is the simple quadratic, and the second and the third term are also differentiable, we can summarize an optimization algorithm for solving the minimization problem (\ref{eq:11}) as follows:

\begin{algorithm} 
    \caption{for solving the minimization problem (\ref{eq:11}) } \label{alg1}
    \begin{algorithmic}
        \STATE \textbf{Inputs:} $y$,\, $\{F_m\}_{m=1}^M$,\, $\alpha$,\, $\beta$,\, $\lambda_1, \lambda_2$, $W$, and max-iter $L$. \vspace{1mm}
        \STATE \textbf{Initialization:} $x^{(0)}=y$,\, $h^{(0)} = h_0$,\, $\gamma_{m,i}^{(0)}=1$,\, $l=0$. \vspace{1mm}
        \STATE \textbf{Iteration:} \vspace{1mm}
        \STATE 1. Obtain convolution matrix $H$ from $h^{(l)}$. \vspace{1mm}
        \STATE 2. $ g_m^{(l)} = F_m x^{(l)}$. \vspace{1mm}
        \STATE 3. Compute $\Lambda(g_m^{(l)})$ according to (\ref{eq:16}). \vspace{1mm}
        \STATE 4. $\gamma_{m,i}^{(l+1)} = (\alpha + 1/2) / (\beta + (1/2) ( g_{m,i}^{(l)} )^2) $.
        \STATE 5. $x^{(l+1)} = (H^T H$ $+ \sum_{m=1}^M F_m^T (\lambda_1 * \text{diag} \{\gamma^{(l)}_m\} $
        \STATE $ \hspace{44 mm} + \lambda_2 * \Lambda(g_m^{(l)} )) F_m)^{-1} H^T y$.
        \STATE 6. Obtain convolution matrix $X$ from $x^{(l)}$. \vspace{1mm}
        \STATE 7. $h^{(l+1)} = (X^TX)^{-1} X^T y$. \vspace{1mm}
        \STATE 8. $l = l+1$. \vspace{1mm}
        \STATE \textbf{Until} $l < L$. \vspace{1mm}
        \STATE \textbf{Output:} $x^{(L)}, h^{(L)}$.
    \end{algorithmic}
\end{algorithm} 

\begin{table*}

    \center

    \resizebox{\textwidth}{!}{
        
        \begin{tabular}{| l | cc | cc | cc | cc | cc | cc | cc | cc |}
        \hline 
         & \multicolumn{2}{c |}{ker01} & \multicolumn{2}{c |}{ker02} & \multicolumn{2}{c |}{ker03} & \multicolumn{2}{c |}{ker04} & \multicolumn{2}{c |}{ker05} & \multicolumn{2}{c |}{ker06} & \multicolumn{2}{c |}{ker07} & \multicolumn{2}{c |  }{ker08} \\

                \hline 
        

         
         \hline
         
         Img01 & \multicolumn{2}{c |}{\enspace42.49 \, \enspace22.85} & \multicolumn{2}{c |}{\enspace43.95 \enspace \textbf{\enspace24.87}} & \multicolumn{2}{c |}{\enspace31.50 \, \textbf{\enspace16.65}} & \multicolumn{2}{c |}{105.58 \, \enspace51.12} & \multicolumn{2}{c |}{\enspace26.62 \, \enspace17.12} & \multicolumn{2}{c |}{\enspace28.29 \, \textbf{\enspace17.82}} & \multicolumn{2}{c |}{\enspace49.07 \, \textbf{\enspace20.87}} & \multicolumn{2}{c |}{\enspace58.55 \, \textbf{\enspace26.14}} \\
         
         & \multicolumn{2}{c |}{\enspace26.07 \, \textbf{\enspace25.45} } & \multicolumn{2}{c |}{\enspace33.21 \, \enspace31.08} & \multicolumn{2}{c |}{\enspace17.07 \, \enspace17.59} & \multicolumn{2}{c |}{\enspace60.31 \, \textbf{\enspace46.52}} & \multicolumn{2}{c |}{\enspace17.37 \, \textbf{\enspace14.39}} & \multicolumn{2}{c |}{\enspace29.91 \, \enspace17.95} & \multicolumn{2}{c |}{\enspace32.49 \, \enspace30.54} & \multicolumn{2}{c |}{\enspace41.06 \, \enspace37.99} \\
         
         \hline 
        
         Img02 & \multicolumn{2}{c |}{\enspace55.61 \, \enspace49.42} & \multicolumn{2}{c |}{\enspace60.51 \, \enspace54.6} & \multicolumn{2}{c |}{\enspace49.98 \, \enspace36.06} & \multicolumn{2}{c |}{102.26 \, \enspace74.03} & \multicolumn{2}{c |}{\enspace29.35 \, \enspace35.42} & \multicolumn{2}{c |}{\enspace29.44 \, \textbf{\enspace20.14}} & \multicolumn{2}{c |}{\enspace57.25 \, \enspace38.23} & \multicolumn{2}{c |}{\enspace69.12 \, \enspace55.23} \\
         
         & \multicolumn{2}{c |}{\enspace42.29 \, \textbf{\enspace34.91}} & \multicolumn{2}{c |}{\textbf{\enspace33.83} \, \enspace35.25} & \multicolumn{2}{c |}{\textbf{\enspace31.66} \, \enspace33.47} & \multicolumn{2}{c |}{\textbf{\enspace45.18} \, \enspace54.54} & \multicolumn{2}{c |}{\enspace38.33 \, \textbf{\enspace23.64}} & \multicolumn{2}{c |}{\enspace76.32 \, \enspace31.36 } & \multicolumn{2}{c |}{\textbf{\enspace32.28} \, \enspace33.82} & \multicolumn{2}{c |}{\textbf{\enspace34.66} \, \enspace37.55} \\    

         \hline 

         Img03 & \multicolumn{2}{c |}{\enspace35.26 \,  \enspace28.66} & \multicolumn{2}{c |}{\enspace42.45 \, \enspace43.96} & \multicolumn{2}{c |}{\enspace17.99 \, \enspace15.36} & \multicolumn{2}{c |}{\enspace67.39 \, \enspace70.82} & \multicolumn{2}{c |}{\enspace17.18 \, \enspace13.94} & \multicolumn{2}{c |}{\enspace21.16 \, \enspace26.75} & \multicolumn{2}{c |}{\enspace27.18 \, \enspace24.35} & \multicolumn{2}{c |}{\enspace37.74 \, \enspace27.83} \\
         
         & \multicolumn{2}{c |}{\enspace19.43 \, \textbf{\enspace19.25}} & \multicolumn{2}{c |}{\textbf{\enspace19.76} \, \enspace23.58} & \multicolumn{2}{c |}{\textbf{\enspace12.90} \, \enspace15.26} & \multicolumn{2}{c |}{\textbf{\enspace25.17} \, \enspace28.26} & \multicolumn{2}{c |}{\textbf{\enspace11.86} \, 12.37} & \multicolumn{2}{c |}{\textbf{\enspace9.65} \, \enspace13.13} & \multicolumn{2}{c |}{\textbf{\enspace11.98} \, \enspace16.59} & \multicolumn{2}{c |}{\textbf{\enspace25.52} \, \enspace27.63} \\
         
         \hline 
         
         Img04 & \multicolumn{2}{c |}{\enspace79.76 \,  \enspace75.56} & \multicolumn{2}{c |}{136.58 \, 201.12} & \multicolumn{2}{c |}{\enspace52.11 \, \enspace24.01} & \multicolumn{2}{c |}{\enspace97.35 \, 261.48} & \multicolumn{2}{c |}{\enspace38.20 \, \enspace23.05} & \multicolumn{2}{c |}{\enspace76.19 \, \enspace65.30} & \multicolumn{2}{c |}{101.42 \, 120.18} & \multicolumn{2}{c |}{118.15 \, 424.02} \\
         
         & \multicolumn{2}{c |}{\enspace45.68 \, \textbf{\enspace42.27}} & \multicolumn{2}{c |}{\enspace75.88 \, \textbf{\enspace71.99}} & \multicolumn{2}{c |}{\textbf{\enspace19.51} \, \enspace22.43} & \multicolumn{2}{c |}{126.19 \, \textbf{\enspace52.47}} & \multicolumn{2}{c |}{\textbf{\enspace17.39} \, \enspace20.27} & \multicolumn{2}{c |}{\enspace39.69 \, \textbf{\enspace30.62}} & \multicolumn{2}{c |}{ \textbf{\enspace54.55} \,  \enspace56.94} & \multicolumn{2}{c |}{\enspace58.35 \, \textbf{\enspace57.95}}\\
       
        \hline 

        \end{tabular}

    }
    \caption{SSD error of 32 test images, achieved by (Top Left) Levin et al. \cite{levin:2011} , (Top Right) Babacan et al. \cite{babacan:2012} , (Bottom Left) Perrone et al. \cite{perrone:2015}, and (Bottom Right) the proposed method  with the same non-blind deblurring algorithm \cite{levin:2007}.}

\end{table*}

\begin{figure}[htb]
 \centering
  \centerline{\includegraphics[width=8.5cm]{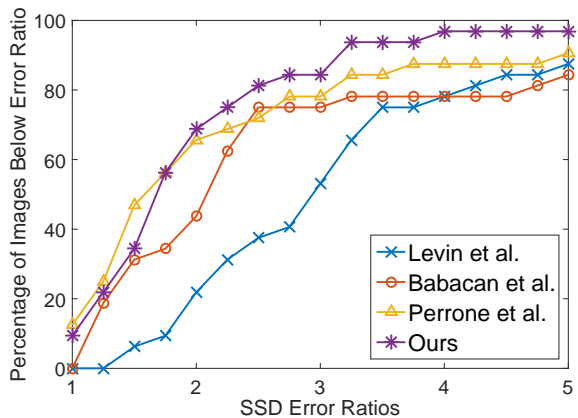}}
\caption{ Cumulative histogram of SSD error ratio }

\end{figure}

\section{Experiment}

We evaluated the proposed algorithms and the other state-of-the-art algorithms \cite{levin:2011,babacan:2012,perrone:2015} on the dataset from Levin et al. \cite{levin:2009}. The dataset is made of 4 images of size $255 \times 255$ pixels blurred with 8 different blur kernels, and it is provided with ground truth sharp images and blur kernels.

In practice, we employed a multiscale approach to deal with the large blur support problem  \cite{fergus:2006}. The input image and the blur are down sampled at each level by $\sqrt{2}$, and the parameter, $\lambda_1$ and $\lambda_2$, are divided by the number 2. Then the algorithm \ref{alg1} was applied at each scale. The number of levels of the pyramid is computed such that at the top level the blur kernel has a support of 3 pixels. We used the fixed parameter values for all the tests, $\lambda_1 =4.5\mathrm{e}{-5}, \lambda_2 = 5\mathrm{e}{-6}, \alpha = 1\mathrm{e}{-18}, \beta = 1/1700$, $W=3$, and 4500 iterations for each pyramid level. The parameter values have been found experimentally. For the other algorithms, we used the parameters provided by the authors.

\begin{figure}[htb]
 \centering
  \centerline{\includegraphics[width=8.5cm]{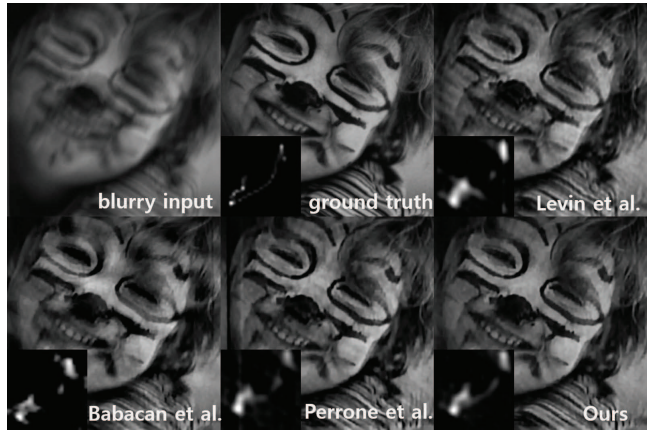}}
\caption{ Blind deconvolution results on Img04 with ker04.}
\end{figure}

First, we measured the sum of squared distance (SSD) between the recovered images and the ground truth images in Table. 1. To measure the effectiveness of estimated blur kernels, the SSD ratio proposed in \cite{levin:2009} was computed. The SSD ratio is defined by $\sum_{i=1}^N ( x^{L}_i - x^{G}_i)^2 /\sum_{i=1}^N (x^{H}_i - x^{G}_i)^2$, where $x^G$ is the ground truth image, $x^L$ is the image obtained by solving a non-blind deconvolution with the estimated blur, and $x^H$ is the image obtained by solving a non-blind deconvolution with the ground truth blur. For all the tests, we used the non-blind deconvolution algorithm from Levin et. al \cite{levin:2007} with $\lambda = 10^{-3}$. 

In Fig.1, we plot the cumulative histogram of SSD ratios (e.g., bin=3 counts the percentage of test examples achieving SSD ratio below 3). Our algorithm performs SSD ratio equal to 2 for more than 60\% of the images, clearly outperforming the method from Levin et al. \cite{levin:2011} and Babacan et al. \cite{babacan:2012}. Our method is on par with high performing blind image deconvolution algorithm (logMM) from Perrone et al. \cite{perrone:2015}. Lastly, Fig. 2 presents some of the estimated images and blur kernels from the experiment. 

\section{Conclusion}

In this paper, we presented a blind image deconvolution algorithm employing structural information among the sparse coefficients. Specifically, a novel problem formulation combining Student's-t image prior and overlapping group sparsity is proposed. Its effectiveness has been demonstrated by the experiment. Future work may include faster approximation of the algorithm using ADMM \cite{boyd:2011} and extensive analysis on the effect of the group size in OGS term.  




\begin{thebibliography}{10}

\bibitem{ill1}
N.~B. Karayiannis and A.~N. Venetsanopoulos,
\newblock ``Regularization theory in image restoration-the stabilizing
  functional approach,''
\newblock {\em IEEE Trans. Acoustic Speech Signal Processing.}, vol. 38, pp.
  1155--1179, 1990.

\bibitem{ill2}
A.~Tikhonov and V.~Arsenin,
\newblock {\em Solution of ill-poised problems},
\newblock Winston, Washington DC, 1977.

\bibitem{natural1}
B.A. Olshausen and D.J. Field,
\newblock ``Emergence of simple-cell receptive field properties by learning a
  sparse code for natural images,''
\newblock {\em Nature}, vol. 381, pp. 607--608, 1996.

\bibitem{natural2}
S.G. Mallat,
\newblock ``A theory for multiresolution signal decomposition: the wavelet
  representation,''
\newblock {\em IEEE Trans. Pattern Analysis and Machine Intelligence.}, vol.
  11, pp. 674--694, 1989.

\bibitem{fergus:2006}
R.~Fergus, B.~Singh, A.~Hertzmann, S.~T. Roweis, and W.~T. Freeman,
\newblock ``Removing camera shake from a single photograph,''
\newblock {\em ACM Trans. Graphics.}, vol. 25, pp. 787--794, 2006.

\bibitem{shan:2008}
Q.~Shan, J.~Jia, and A.~Agarwala,
\newblock ``High-quality motion deblurring from a single image,''
\newblock {\em ACM Trans. Graphics.}, 2008.

\bibitem{cho:2009}
S.~cho and S.~Lee,
\newblock ``Fast motion deblurring,''
\newblock {\em ACM Trans. Graphics.}, vol. 28, 2009.

\bibitem{ker:2009}
D.~Krishnan and R.~Fergus,
\newblock ``Fast image deconvolution using hyper-laplacian priors,''
\newblock in {\em NIPS}, 2009.

\bibitem{dim:2009}
D.~G. Tzikas, A.~C. Kikas, and N.~P. Galatsanos,
\newblock ``Variational bayesian kernel-based blind image deconvolution with
  student's-t ptiors,''
\newblock {\em IEEE Trans. Image Processing}, vol. 18, pp. 753--764, 2009.

\bibitem{levin:2009}
A.~Levin, Y.~Weiss, F.~Durand, and W.~T. Freeman,
\newblock ``Understanging and evaluating blind deconvolution algorithms,''
\newblock in {\em CVPR}, 2009.

\bibitem{levin:2011}
A.~Levin, Y.~Weiss, F.~Durand, and W.~T. Freeman,
\newblock ``Efficient marginal likelihood optimization in blind
  deconvolution,''
\newblock in {\em CVPR}, 2011.

\bibitem{ker:2011}
D.~Krishnan, T.~Tay, and R.~Fergus,
\newblock ``Blind deconvolution using a normalized sparsity measure,''
\newblock in {\em CVPR}, 2011.

\bibitem{wipf:2014}
D.~Wipf and H.~Zhang,
\newblock ``Revisiting bayesian blind deconvolution,''
\newblock {\em Journal of Machine Learning Research}, vol. 15, pp. 3595--3634,
  2014.

\bibitem{babacan:2012}
S.~D. Babacan, R.~Molina, and M.~N. Do,
\newblock ``Bayesian blind deconvolution with general sparse image priors,''
\newblock in {\em ECCV}, 2012.

\bibitem{perrone:2015}
D.~Perrone and P.~Favaro,
\newblock ``A logarithmic image prior for blind deconvolution,''
\newblock {\em Int. J. Comput Vis.}, 2015.

\bibitem{peyr:2011}
G.~Peyre and J.~Fadili,
\newblock ``Group sparsity with overlapping partition functions,''
\newblock in {\em In Proc. European Sig. Image Proc. Conf. (EUSIPCO)}, 2011.

\bibitem{figu:2011}
M.~Figueiredo and J.~Bioucas-Dias,
\newblock ``An alternating direction algorithm for (overlapping) group
  regularization,''
\newblock in {\em In Signal Process. Adaptive Sparse Structured Represntation},
  2011.

\bibitem{bach:2012}
F.~Bach, R.~Jenatton, J.~Mairl, and G.~Obozinski,
\newblock ``Structured sparsity through convex optimization,''
\newblock {\em Stat. Sci.}, vol. 27, pp. 450--468, 2012.

\bibitem{sele:2013}
I.W. Selesnik and P.Y. Chen,
\newblock ``Total variation denoising with overlapping group sparsity,''
\newblock in {\em Proc. IEEE Int. Conf. Acoust., Speech Signal Process.}, 2013.

\bibitem{liu:2015}
J.~Liu, T.~Huang, I.W. Selesnick, X.Lv, and P.~Chen,
\newblock ``Image restoration using total variation with overlapping group
  sparsity,''
\newblock {\em Inf. Sci.}, vol. 295, pp. 232--246, 2015.

\bibitem{shi:2016}
M.~Shi, T.~Han, and S.~Liu,
\newblock ``Total variation image restoration using hyper-lapacian prior with
  overlapping group sparsity,''
\newblock {\em Signal Processing}, vol. 126, pp. 65--76, 2016.

\bibitem{sele:2015}
Ivan~W. Selesnick,
\newblock ``Generalized total variation,''
\newblock in {\em Signal Processing Letters}. IEEE, 2015, vol.~22.

\bibitem{levin:2007}
A.~Levin, R.~Fergus, F.~Durand, and W.~T. Freeman,
\newblock ``Image and depth from a conventional camera with a coded aperture,''
\newblock in {\em Trans. Graph.} ACM, 2007, vol. 26(3).

\bibitem{boyd:2011}
S.~Boyd, N.~Parikh, E.~Chu, B.~Peleato, and J.~Eckstein,
\newblock {\em Distributed Optimization and Statistical Learning via the
  Alternating Direction Method of Multipliers},
\newblock Foundations and Trends® in Machine Learning, Now Publishers Inc.
  Hanover, MA, USA, 1st edition, 2011.

\end{thebibliography}

\end{document}